\documentclass[conference, 10pt]{IEEEtran}
\usepackage{epsfig,endnotes}
\usepackage[hidelinks]{hyperref}
\usepackage{url}
\usepackage{xcolor,colortbl}
\usepackage{booktabs,caption,fixltx2e}
\usepackage[flushleft]{threeparttable}
\usepackage{wrapfig}
\usepackage{makecell}
\usepackage{float,graphicx}
\usepackage[font=footnotesize]{caption}
\usepackage{subcaption}
\usepackage{multicol}
\usepackage{multirow}
\usepackage{float}
\usepackage{tikz}
\usepackage{mwe}
\usepackage{listings}
\usepackage{amsmath,mathtools,amssymb}
\usepackage{array, makecell}
\usepackage{enumitem}
\usepackage{microtype}
\usepackage{verbatim}
\usepackage{setspace}
\usepackage{bookmark}
\usepackage[ruled,linesnumbered]{algorithm2e}

\SetCommentSty{mycommfont}
\usepackage{xcolor,colortbl}
\definecolor{ashgrey}{rgb}{0.7, 0.75, 0.71}
\SetKw{KwBy}{by}
\usepackage{textcomp}
\makeatletter
\g@addto@macro{\UrlBreaks}{\UrlOrds}

\definecolor{Gray}{gray}{0.85}
\definecolor{lblue}{rgb}{0.7,1,1}
\usepackage[belowskip=-15pt,aboveskip=0pt]{caption}

\definecolor{codegray}{rgb}{0.5,0.5,0.5}
\definecolor{codepurple}{rgb}{0.58,0,0.82}
\definecolor{backcolour}{rgb}{0.95,0.95,0.92}

\lstdefinestyle{mystyle}{
    backgroundcolor=\color{backcolour},   
    commentstyle=\color{codegray},
    keywordstyle=\color{magenta},
    numberstyle=\tiny\color{codegray},
    stringstyle=\color{codepurple},
    basicstyle=\ttfamily\scriptsize,
    breakatwhitespace=false,         
    breaklines=true,                 
    captionpos=b,                    
    keepspaces=true,                               
    numbersep=5pt,                  
    showspaces=false,                
    showstringspaces=false,
    showtabs=false,                  
    tabsize=2
}

\def\endthebibliography{%
  \def\@noitemerr{\@latex@warning{Empty `thebibliography' environment}}%
  \endlist
}
\ifCLASSINFOpdf
\else
\fi

\begin{document}

\title{Learning Structural Manipulability in Gate-Level Netlists Using Graph Neural Networks}

\author{
  \IEEEauthorblockN{Rupesh Raj Karn, Ozgur Sinanoglu}\vspace{-10pt}\\
  \IEEEauthorblockA{Center for Cyber Security, New York University, Abu Dhabi, UAE.}\vspace{-10pt}\\
  Email: \{rupesh.k, ozgursin\}@nyu.edu
}


\maketitle

\begin{abstract}
Gate-level netlists exhibit intrinsic structural properties that influence signal propagation independently of functional simulation. We define a topology-driven structural manipulability score that characterizes node-level structural flexibility using path participation, k-core embedding, symmetry, and centrality. Modeling netlists as directed graphs, we formulate node-level regression to learn this topology-derived score using graph neural networks (GNNs). Experiments on ISCAS85 and EPFL benchmarks evaluate how effectively different GNN architectures approximate this metric across held-out circuits, with hierarchical models yielding the most consistent rankings. Component-level and ablation analyses examine the contribution of individual factors. As an illustrative case study, analysis of Trojan-injected circuits using TrustHub templates reveals statistically distinguishable structural patterns, indicating that topology-based scoring provides complementary structural insight.
\end{abstract}

\IEEEpeerreviewmaketitle

\begin{IEEEkeywords}
Structural Topology, Circuit Structural Criticality, Gate-level Netlists, Graph Neural Networks, Embeddings
\end{IEEEkeywords}

\vspace{-10pt}
\section{Introduction}
\label{sec:introduction}

Gate-level netlists are traditionally analyzed using functional simulation, formal verification, and timing analysis to ensure correctness and performance \cite{guo2023general, ain2024formal}. While indispensable, these methods primarily capture \emph{functional} behavior and do not explicitly characterize intrinsic \emph{structural topology} \cite{zhao2023hybridnet} independent of input stimuli. As designs scale, understanding graph-level structural organization becomes important for complementary analysis of connectivity, robustness, and embedding patterns. 

Structural topology \cite{zhao2023hybridnet} governs information flow, embedding density, and how local changes propagate through the circuit graph. Unlike simulation, topology-driven analysis is input-agnostic and captures global properties such as centrality, redundancy, symmetry, and embedding depth directly from the netlist. It also avoids the computational cost of exhaustive input evaluation \cite{guo2023general}.

In this work, we define \emph{structural manipulability} as a node-level score derived from graph-theoretic properties of gate-level netlists represented as directed graphs. Here, ``manipulability'' denotes relative structural flexibility inferred from connectivity patterns, not a validated measure of engineering effort, edit feasibility, timing impact, functional correctness, or attack success. Accordingly, the score is a \emph{designed structural proxy}: its values depend on the chosen netlist representation and may change under logic-preserving transformations. Our goal is therefore not to establish ground-truth manipulability, but to define a consistent topology-derived target for structural characterization and GNN-based approximation.

To approximate this score, we employ graph neural networks (GNNs) \cite{alrahis2022embracing}, which aggregate local neighborhood information to capture global structure. The task is formulated as node-level regression, where GNNs learn to approximate the predefined topology-derived score.

In summary, the main contributions of this work are:
\begin{enumerate}
    \item We formalize a topology-driven structural manipulability score based on path participation, core embedding, symmetry, and centrality.
    \item We evaluate how effectively GNNs approximate this score across circuits from ISCAS85 and EPFL benchmarks.
    \item We provide component-level and ablation analyses to examine the contribution of individual structural factors.
    \item We analyze per-circuit behavior and architectural biases to characterize variability across netlist families.
    \item We present a preliminary case study on Trojan-injected circuits using synthetic TrustHub-style insertions, illustrating statistically distinguishable structural patterns.
\end{enumerate}

We have released the source code at \cite{gnnmutability2026}.

\section{Prior-art Comparison}
\label{sec:prior-art}

Recent work has applied GNNs to hardware security and EDA problems, including hardware Trojan detection, timing prediction, and logic optimization. For example, GNN-based Trojan detection frameworks perform supervised node- or graph-level classification on netlist graphs \cite{yasaei2022hardware}, while graph learning has been used for delay and performance estimation in EDA flows \cite{zhang2020circuitgnn, chen2021gnn_timing}. Classical graph-theoretic analyses employ centrality and $k$-core measures to study structural properties \cite{kitsak2010identification, freeman1977centrality}, and Weisfeiler--Lehman (WL) hashing has been used for structural matching and IP protection \cite{shervashidze2011weisfeiler}. 

These approaches fall into three categories: supervised learning for specific tasks, performance-oriented prediction, and direct computation of individual structural metrics. In contrast, our work defines a composite topology-driven structural score that integrates multiple graph-theoretic properties into a unified node-level quantity, providing a consistent characterization of embedding, redundancy, and connectivity. Methodologically, we differ from direct metric computation by studying whether this composite score can be approximated using GNNs. This frames the problem as learning a surrogate for a topology-derived quantity from graph structure, focusing on representational learnability rather than task-specific prediction.

Finally, we do not claim superiority over existing EDA tools or security methods. The proposed score is intended only as a topology-based structural characterization primitive, not as a substitute for established functional, timing, or security analyses.

\section{Structural Manipulability Metric}
\label{sec:manipulability_metric}

A gate-level netlist is modeled as a directed graph 
$G = (V, E)$, where nodes represent logic gates (or sequential elements) and edges represent signal propagation. Let $A$ denote the adjacency matrix, $n = |V|$, $m = |E|$, and $\mathcal{P}$ the set of primary inputs (PI) to primary outputs (PO) paths. For each node $v$, $\deg^{+}(v)$ and $\deg^{-}(v)$ denote fan-out and fan-in, respectively. All metrics are computed directly from $G$ without functional simulation.

\vspace{-5pt}
\subsection{Definition of Structural Manipulability}
\label{subsec:manipulatively_definition}

We define the \emph{structural manipulability score} $M(v)$ as a composite of graph-theoretic properties capturing structural flexibility. The score is normalized to $[0,1]$, and we use uniform weights $\alpha = \beta = \gamma = \delta$.

\textbf{Path Participation Ratio.} Let $\pi(v)$ denote the fraction of PI-to-PO paths traversing node $v$:
\begin{equation}
    \pi(v) = \frac{|\{ p \in \mathcal{P} : v \in p \}|}{|\mathcal{P}|}, 
    \qquad
    M_{\text{path}}(v) = 1 - \pi(v).
\end{equation}
$\pi(v)$ is approximated via path sampling or normalized betweenness centrality.

\textbf{Core Decomposition Score.} Let $k(v)$ denote the core number from $k$-core decomposition \cite{kitsak2010identification}:
\begin{equation}
    M_{\text{core}}(v) = \frac{k(v)}{\max_{u \in V} k(u)}.
\end{equation}
This captures structural embedding depth.

\textbf{Symmetry Class Score.} Let $\phi(v)$ denote the Weisfeiler--Lehman (WL) hash \cite{shervashidze2011weisfeiler}, with class
\begin{equation}
    \mathcal{C}(v) = \{ u \in V : \phi(u) = \phi(v) \},
\end{equation}
and
\begin{equation}
    M_{\text{sym}}(v) = \frac{|\mathcal{C}(v)|}{n}.
\end{equation}

\textbf{Centrality-Based Score.} Let $b(v)$ denote normalized betweenness centrality \cite{freeman1977centrality}:
\begin{equation}
    M_{\text{cent}}(v) = 1 - b(v).
\end{equation}
While both $M_{\text{path}}(v)$ and $M_{\text{cent}}(v)$ rely on global flow-related quantities, they capture overlapping aspects of structural criticality. In particular, betweenness centrality can be viewed as an aggregate proxy for path participation, leading to partial redundancy between these components. We retain both terms to provide complementary perspectives on global connectivity: $M_{\text{path}}(v)$ reflects explicit path coverage, whereas $M_{\text{cent}}(v)$ captures flow concentration through bottleneck structures.

\textbf{Composite Structural Manipulability.} The final score is:
\begin{equation}
    M(v) = \alpha M_{\text{path}}(v)
      + \beta M_{\text{core}}(v)
      + \gamma M_{\text{sym}}(v)
      + \delta M_{\text{cent}}(v),
\end{equation}
with $\alpha + \beta + \gamma + \delta = 1$. 

$M(v)$ reflects standard structural properties: nodes with low path participation or centrality tend to have higher scores, while higher $k$-core values indicate stronger embedding. Empirically, $M_{\text{core}}(v)$ dominates variation in $M(v)$, with other components providing secondary refinement.

Collectively, $M(v)$ captures structural flexibility using intrinsic connectivity patterns.

\vspace{-5pt}
\subsection{Learning Structural Manipulability with GNNs}
\label{subsec:manipulability_gnns}

Given node features $X \in \mathbb{R}^{n \times d}$ and graph structure $A$, a GNN learns node representations via message passing:
\begin{equation}
    H^{(l+1)} = \sigma \left( \mathcal{A}(A) H^{(l)} W^{(l)} \right),
\end{equation}
where $\mathcal{A}(A)$ is a normalized aggregation operator, $W^{(l)}$ are learnable weights, and $\sigma$ is a non-linear activation \cite{wu2020comprehensive}.

Although the constituent graph metrics can be computed directly, repeated evaluation of global structural quantities (such as centrality or path-based measures) can be computationally expensive or cumbersome across large collections of circuits and design variants. We therefore investigate whether a learned model can approximate the composite score directly from graph structure. This formulation treats the problem as learning a surrogate for a topology-derived quantity, rather than discovering an externally validated ground-truth property.

The prediction task is formulated as node-level regression:
\begin{equation}
    \hat{M}(v) = f_{\theta}(G, v),
\end{equation}
trained using mean squared error:
\begin{equation}
    \mathcal{L} = \frac{1}{n} \sum_{v \in V} \| \hat{M}(v) - M(v) \|_2^2.
\end{equation}
This formulation treats structural manipulability as a topology-driven target that is independent of the learning architecture. Different GNN models do not alter the definition of $M(v)$; instead, they differ in how effectively they approximate this predefined structural score through localized message passing.

As an illustrative example, a GCN (graph convolution network) aggregates normalized neighborhood features and performs diffusion-like smoothing over the netlist graph, which favors stable estimation of global structural quantities such as core embedding and centrality. Deeper propagation enables approximation of the dominant components of $M(v)$, although excessive smoothing may reduce contrast across nodes. Architectures such as GSAGE (graph sample and aggregate) and GAT (graph attention network) modify the aggregation mechanism through inductive sampling or attention weighting, influencing the dynamic range and emphasis of structural signals. Graph isomorphism networks (GIN), while highly expressive under the 1-WL (1-dimensional Weisfeiler–Lehman test \cite{shervashidze2011weisfeiler}) framework, enhance discrimination of locally distinct structural neighborhoods. However, since $M(v)$ integrates both local and global structural properties, architectural variations do not alter its definition but influence how effectively global structural embedding is approximated through localized message passing. Similar considerations apply across other GNN variants \cite{wu2020comprehensive}.

\vspace{-10pt}
\section{Experiments}
\label{sec:experiments}

The experiments evaluate the \emph{learnability} and consistency of the structural manipulability score $M(v)$ across circuits and architectures. The goal is not downstream task performance, but to assess how effectively topology-driven structural properties can be approximated from graph representations.

\subsection{Experiment Testbench}
\label{subsec:expt_testbench}

Experiments are conducted on gate-level netlists from the ISCAS85 and EPFL benchmark suites\footnote{\url{https://github.com/jpsety/verilog_benchmark_circuits}}. These synthesized benchmarks are widely used for structural analysis. The framework operates on gate-level graph representations independent of HDL origin. We evaluate a diverse set of GNN architectures including GIN, GCN, GraphSAGE, GAT, message passing neural networks (MPNN), approximate personalized propagation of neural predictions (APPNP), heterogeneous graph neural networks (HetGNN), graph U-Net (g-U-Net), signed graph neural networks (SGNN), and graph transformer networks (GTN). These architectures are explained in \cite{wu2020comprehensive}.

\vspace{-5pt}
\subsection{Comparative Analysis of GNN Architectures}
\label{subsec:gnn_comparison}

Table~\ref{tab:gnn_comparison} reports node-level regression performance for predicting $M(v)$ on held-out circuits. All models use identical splits and normalization. Multiple circuit-level splits are used, where different subsets of circuits are reserved for testing across seeds.

Most spatial message-passing models (GCN, GSAGE, MPNN) achieve low MSE ($\approx 3\times 10^{-4}$ to $5\times 10^{-4}$) and relatively strong rank correlation (Spearman $\approx 0.75$--$0.79$), indicating that the topology-derived score is largely recoverable from local aggregation. GCN performs competitively, suggesting diffusion-based propagation captures key structural signals.

GAT and GTN show comparable performance, indicating attention-based weighting refines but does not fundamentally change prediction behavior. Graph U-Net achieves the highest Spearman correlation, suggesting benefits from hierarchical representations.

GIN shows slightly lower rank correlation, possibly due to reduced bias toward smooth structural metrics. APPNP remains stable, while HetGNN indicates moderate gains from node-type awareness. SGNN significantly underperforms, suggesting signed edge assumptions are not aligned with netlist structure. These results are consistent with the hypothesis outlined in Section~\ref{subsec:manipulability_gnns}, suggesting that different GNN architectures vary in their ability to approximate the topology-derived score $M(v)$.

Overall, these results show that the proposed structural manipulability score can be approximated by a range of GNN architectures, with architectural inductive biases influencing ranking stability across circuits. However, as discussed in subsequent sections, per-circuit variability and benchmark-specific effects indicate that this approximation is not uniformly consistent across all designs, and cross-design transfer should be interpreted within the scope of the evaluated benchmark setting.

\begin{table}[!t]
\centering
\vspace{10pt}
\caption{Node-level prediction performance of different GNN architectures on the structural manipulability score $M(v)$. Results are reported as mean $\pm$ standard deviation over different seeds. MSE denotes mean squared error, MAE denotes mean absolute error, and Spearman denotes the Spearman rank correlation coefficient \cite{yun2021neo} between predicted $\hat{M}(v)$ and ground-truth $M(v)$.}
\label{tab:gnn_comparison}
\scriptsize
\begin{tabular}{lccc}
\toprule
\textbf{Model} & \textbf{MSE} & \textbf{MAE} & \textbf{Spearman} \\
\midrule
GCN              & $0.00031 \pm 0.00005$ & $0.01152 \pm 0.00127$ & $0.7881 \pm 0.0126$ \\
GSAGE            & $0.00044 \pm 0.00004$ & $0.01359 \pm 0.00169$ & $0.7569 \pm 0.0372$ \\
GIN              & $0.00046 \pm 0.00009$ & $0.01375 \pm 0.00165$ & $0.6807 \pm 0.0056$ \\
GAT              & $0.00045 \pm 0.00005$ & $0.01319 \pm 0.00129$ & $0.7430 \pm 0.0311$ \\
MPNN             & $0.00048 \pm 0.00007$ & $0.01348 \pm 0.00143$ & $0.7600 \pm 0.0166$ \\
APPNP            & $0.00043 \pm 0.00007$ & $0.01284 \pm 0.00134$ & $0.7250 \pm 0.0141$ \\
G-U-Net          & $0.00045 \pm 0.00006$ & $0.01344 \pm 0.00119$ & $0.8129 \pm 0.0205$ \\
HetGNN           & $0.00050 \pm 0.00005$ & $0.01372 \pm 0.00136$ & $0.7706 \pm 0.0236$ \\
SGNN             & $0.20858 \pm 0.01203$ & $0.15210 \pm 0.00584$ & $0.4894 \pm 0.2005$ \\
GTN              & $0.00050 \pm 0.00010$ & $0.01329 \pm 0.00186$ & $0.7952 \pm 0.0301$ \\
\bottomrule
\end{tabular}
\end{table}

\begin{table}[!t]
\centering
\vspace{10pt}
\caption{Node-level stats of structural manipulability components and their Spearman rank correlation with the composite score $M(v)$.}
\label{tab:component_stats}
\small
\begin{tabular}{lccc}
\toprule
\textbf{Component} & \textbf{Mean} & \textbf{Std} & \textbf{Spearman$(M_{\text{comp}}(v),\, M(v))$} \\
\midrule
$M_{\text{path}}$ & 0.992 & 0.044 & 0.144 \\
$M_{\text{core}}$ & 0.848 & 0.193 & 0.763 \\
$M_{\text{sym}}$  & 0.046 & 0.079 & 0.468 \\
$M_{\text{cent}}$ & 0.336 & 0.124 & 0.249 \\
\bottomrule
\end{tabular}
\vspace{-10pt}
\end{table}

\vspace{-5pt}
\subsection{Component-Level Analysis of Structural Manipulability}
\label{subsec:component_analysis}

To examine the composition of the composite score $M(v)$ (Section~\ref{subsec:manipulatively_definition}), we analyze the distribution and influence of its components in Table~\ref{tab:component_stats}. The core-based term $M_{\text{core}}$ shows the strongest correlation with $M(v)$ (Spearman $=0.763$), indicating that structural embedding depth is the dominant factor. This is consistent with the ablation study (Table~\ref{tab:ablation}, Section~\ref{subsec:ablation}), where removing $M_{\text{core}}$ causes the largest performance drop. Its relatively high variance further suggests that core structure drives most variation across nodes.

The symmetry component $M_{\text{sym}}$ shows moderate correlation with the composite score (Spearman $=0.468$), but its overall mean is low ($0.046$), reflecting that large symmetry classes are relatively rare in these benchmarks. The centrality-based term $M_{\text{cent}}$ has a weaker but non-negligible correlation (Spearman $=0.249$), indicating that structural bottleneck properties contribute secondary refinement to the overall score. Finally, the path-participation proxy $M_{\text{path}}$ exhibits high mean ($0.992$) and very low variance, resulting in minimal correlation with the composite score. This suggests that the approximation used for path participation provides limited discriminative power across nodes in these benchmarks. 

These results indicate that the composite metric is not empirically balanced across its components. In practice, $M(v)$ is largely driven by the core-based term $M_{\text{core}}$, with symmetry and centrality providing auxiliary structural refinement, and path participation contributing minimally under the current approximation. Accordingly, the proposed score can be interpreted as a core-centered structural characterization enriched by additional graph-theoretic signals. 

Importantly, this empirical imbalance does not imply that the observed component ranking will remain the same in all practical settings, nor does it establish that core structure alone is sufficient to capture all aspects of structural manipulability. Rather, it reflects how these components behave under the specific benchmark distributions considered in this study. This observation is consistent with the formulation in Section~\ref{subsec:manipulatively_definition} and helps explain why GNN architectures (of Section~\ref{subsec:manipulability_gnns}) that better capture global structural embedding achieve stronger ranking performance.

\begin{figure}[!t]
	\begin{center}
    \vspace{-10pt}
		\includegraphics[scale=0.65, trim = {0cm 0cm 0cm 0cm}, clip]{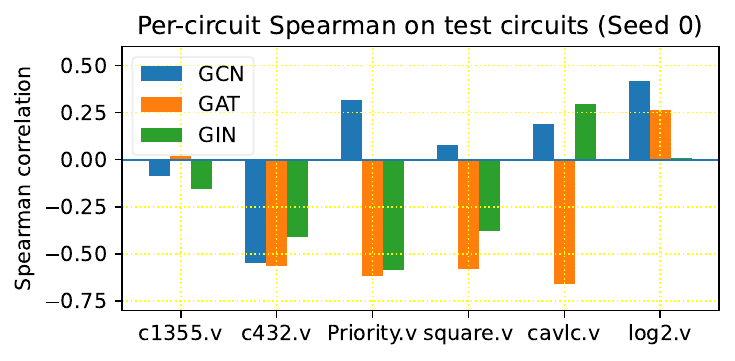} 
        \includegraphics[scale=0.65, trim = {0cm 0cm 0cm 0cm}, clip]{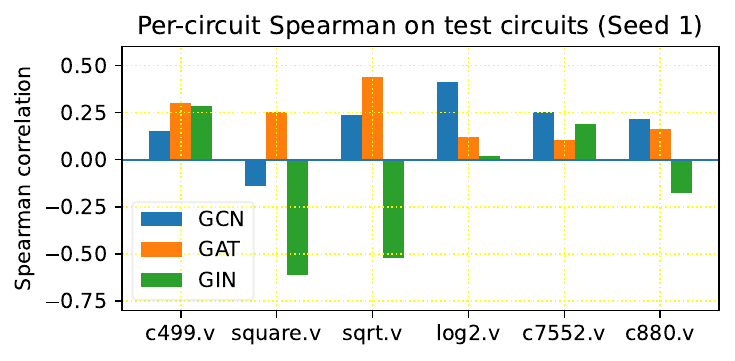} 
		\caption{Per-circuit Spearman analysis for GCN, GAT, and GIN.}
		\label{fig:per_circuit_spearman}
	\end{center}
\end{figure}

\begin{table}[t]
\centering
\vspace{10pt}
\caption{Per-circuit statistics corresponding to Fig.~\ref{fig:per_circuit_spearman}. 
For each seed and model, $\mu_{\text{ISCAS}}$ and $\mu_{\text{EPFL}}$ denote the mean Spearman rank correlation computed over ISCAS and EPFL circuits, respectively. $\rho(|V|,\text{Spr})$ denotes the Pearson correlation between circuit node count $|V|$ and per-circuit Spearman correlation, measuring sensitivity to graph scale. $\Delta_{\text{wt-unwt}}$ represents the difference between size-weighted and unweighted average Spearman values, quantifying the extent to which larger circuits dominate the aggregate ranking performance.}
\label{tab:per_circuit_stats}
\small
\begin{tabular}{llcccc}
\toprule
\textbf{Seed} & \textbf{Model} &
$\boldsymbol{\mu_{\text{ISCAS}}}$ &
$\boldsymbol{\mu_{\text{EPFL}}}$ &
$\boldsymbol{\rho(|V|,\text{Spr})}$ &
$\boldsymbol{\Delta_{\text{wt-unwt}}}$ \\
\midrule
\multirow{3}{*}{Seed 0} 
& GCN & $-0.317$ & $+0.249$ & $+0.484$ & $+0.204$ \\
& GAT & $-0.272$ & $-0.399$ & $+0.490$ & $+0.236$ \\
& GIN & $-0.284$ & $-0.165$ & $+0.111$ & $+0.043$ \\
\midrule
\multirow{3}{*}{Seed 1}
& GCN & $+0.208$ & $+0.170$ & $+0.037$ & $+0.006$ \\
& GAT & $+0.189$ & $+0.270$ & $+0.223$ & $+0.025$ \\
& GIN & $+0.100$ & $-0.372$ & $-0.576$ & $-0.188$ \\
\midrule
\multirow{3}{*}{Pooled}
& GCN & $-0.002$ & $+0.215$ & $+0.336$ & $+0.099$ \\
& GAT & $+0.005$ & $-0.112$ & $+0.363$ & $+0.163$ \\
& GIN & $-0.053$ & $-0.254$ & $-0.240$ & $-0.086$ \\
\bottomrule
\end{tabular}
\vspace{-10pt}
\end{table}

\vspace{-5pt}
\subsection{Per-Circuit Performance Analysis}
\label{subsec:per_circuit_analysis}

Fig.~\ref{fig:per_circuit_spearman} reports circuit-level rank correlation for GCN, GAT, and GIN, highlighting variability not visible in aggregate metrics. Similar results for other GNNs are available in \cite{gnnmutability2026}. Table~\ref{tab:per_circuit_stats} summarizes family-wise means and size-related statistics.

\textbf{Stability across circuit families (ISCAS vs EPFL).}
Performance varies across benchmark families. In Seed~0, GCN shows a strong gap ($\mu_{\text{ISCAS}}=-0.317$ vs.\ $\mu_{\text{EPFL}}=+0.249$), while Seed~1 shows reduced separation ($+0.208$ vs.\ $+0.170$). GIN is highly sensitive, dropping on EPFL circuits in Seed~1 ($\mu_{\text{EPFL}}=-0.372$). Notably, several circuits exhibit negative Spearman correlation, indicating inverted rankings relative to the ground truth. These cases suggest that the learned surrogate is not uniformly reliable across circuit families.

\textbf{Sensitivity to circuit size or structure.}
Circuit size influences ranking stability, but inconsistently. In Seed~0, GCN and GAT show moderate positive correlation with $|V|$ ($\rho\approx 0.48$--$0.49$), while Seed~1 shows weaker or mixed trends (GCN: $\rho=0.037$, GAT: $\rho=0.223$, GIN: $\rho=-0.576$). This variability indicates that size effects depend on structural composition rather than scale alone.

\textbf{Are global results dominated by large circuits?} Because node-level metrics implicitly weight large circuits more, strong global Spearman can be inflated if large designs are easier. This effect is measured via $\Delta_{\text{wt-unwt}}$ in Table~\ref{tab:per_circuit_stats}. For Seed~0, both GCN and GAT have large positive $\Delta_{\text{wt-unwt}}$ ($+0.204$ and $+0.236$), implying that their aggregate Spearman is substantially boosted by large circuits. In contrast, Seed~1 exhibits minimal dominance for GCN ($\Delta=+0.006$) and a modest effect for GAT ($\Delta=+0.025$). GIN shows negative $\Delta$ in Seed~1 ($-0.188$), confirming that its poorer performance on large EPFL circuits pulls down the weighted statistic. The pooled results preserve these trends: GCN and GAT remain positively size-influenced ($\Delta=+0.099$ and $+0.163$), whereas GIN remains negatively influenced ($\Delta=-0.086$).

Overall, Fig.~\ref{fig:per_circuit_spearman} shows that structural manipulability prediction is not uniformly stable across benchmarks. Negative correlations, family-dependent behavior, and size effects indicate that approximation quality varies across circuit types and scales. The learned surrogate for $M(v)$ should therefore be interpreted as capturing structural trends under benchmark conditions rather than providing consistent ranking across all netlists.

\begin{table}[!t]
\centering
\vspace{5pt}
\caption{Ablation study on structural components of $M(v)$ using a representative GCN. Each variant removes one structural component from the composite metric. Results are reported as mean $\pm$ std over two circuit-level splits (two seeds). Other GNNs results are in \cite{gnnmutability2026}.}
\label{tab:ablation}
\small
\begin{tabular}{lccc}
\toprule
\textbf{Target Variant} & \textbf{MSE} & \textbf{MAE} & \textbf{Spearman} \\
\midrule
Full $M(v)$                 & $0.001 \pm 0.000$ & $0.020 \pm 0.004$ & $0.811 \pm 0.006$ \\
$M(v)\setminus M_{\text{sym}}$  & $0.003 \pm 0.001$ & $0.032 \pm 0.005$ & $0.810 \pm 0.001$ \\
$M(v)\setminus M_{\text{core}}$ & $0.000 \pm 0.000$ & $0.003 \pm 0.001$ & $0.554 \pm 0.112$ \\
$M(v)\setminus M_{\text{cent}}$ & $0.021 \pm 0.002$ & $0.095 \pm 0.014$ & $0.690 \pm 0.014$ \\
\bottomrule
\end{tabular}
\vspace{-10pt}
\end{table}

\subsection{Ablation Study on Structural Components of $M(v)$}
\label{subsec:ablation}

To quantify the contribution of each component in $M(v)$, we conduct an ablation study using a representative GCN. The composite score includes path participation, centrality ($M_{\text{cent}}$), core structure ($M_{\text{core}}$), and symmetry ($M_{\text{sym}}$). Each component is removed in turn while re-normalizing the remaining terms. Models are trained under identical circuit-level splits. Table~\ref{tab:ablation} reports the results, with the full metric achieving Spearman $0.811 \pm 0.006$.\footnote{The path-participation term $M_{\text{path}}$ is not included as a separate ablation target. As shown in Table~\ref{tab:component_stats}, it has high mean, low variance, and weak correlation with $M(v)$, indicating limited standalone influence. It also overlaps with $M_{\text{cent}}$ as both capture global flow structure. The ablation therefore focuses on components contributing the most observable variation.}

Removing $M_{\text{sym}}$ has minimal impact ($0.811 \rightarrow 0.810$), though MSE increases slightly. In contrast, removing $M_{\text{core}}$ causes a large drop ($0.811 \rightarrow 0.554$), confirming its dominant role. Removing $M_{\text{cent}}$ also reduces performance ($0.811 \rightarrow 0.690$) and increases error, indicating the importance of global connectivity.

Overall, $M_{\text{core}}$ and $M_{\text{cent}}$ dominate learnability, while symmetry provides secondary refinement. The exclusion of $M_{\text{path}}$ reflects its low variance and redundancy, not omission of a negative result. These findings indicate that $M(v)$ is primarily driven by global embedding and flow structure, with other components providing auxiliary refinement under the evaluated benchmark setting.

\begin{table}[t]
\centering
\vspace{10pt}
\caption{Correlation between circuit-level statistics and per-circuit Spearman correlation computed across all circuits of ISCAS85 and EPFL benchmarks. Only GCN is used in this part of the experiment.}
\label{tab:global_stat_correlation}
\small
\begin{tabular}{lc}
\toprule
\textbf{Statistic} & \textbf{Correlation with Spearman} \\
\midrule
Number of nodes ($|V|$)        & $-0.166$ \\
Number of edges ($|E|$)        & $-0.162$ \\
Average fan-in                 & $+0.273$ \\
Average fan-out                & $+0.273$ \\
\bottomrule
\end{tabular}
\vspace{-10pt}
\end{table}

\vspace{-5pt}
\subsection{Correlation with Circuit Structural Statistics}
\label{subsec:global_stat_correlation}

To examine structural factors affecting learnability of $M(v)$, we correlate per-circuit Spearman performance with global graph statistics across all circuits of ISCAS85 and EPFL benchmarks. A representative GCN model is trained on the full dataset, and the resulting circuit-level Spearman correlations are analyzed against circuit size and structural density metrics.

As shown in Table~\ref{tab:global_stat_correlation}, ranking performance exhibits weak negative correlation with the number of nodes ($\rho = -0.166$) and edges ($\rho = -0.162$), indicating that larger circuits do not inherently improve structural predictability under the evaluated setting. In contrast, average fan-in and fan-out show moderate positive correlation ($\rho = +0.273$), suggesting that circuits with higher local connectivity density provide stronger structural signals for approximating $M(v)$. These results indicate that structural learnability is more closely associated with connectivity density than with absolute graph size. However, the relatively weak magnitude of these correlations suggests that global graph statistics alone do not fully explain per-circuit variability, and other structural factors contribute to the observed differences in prediction quality.

\vspace{-5pt}
\subsection{Robustness Across Design Styles and Circuit Sizes} \label{subsec:robustness} 

\begin{figure}[!t]
\vspace{-10pt}
	\begin{center}
		\includegraphics[scale=0.6, trim = {0cm 0cm 0cm 0cm}, clip]{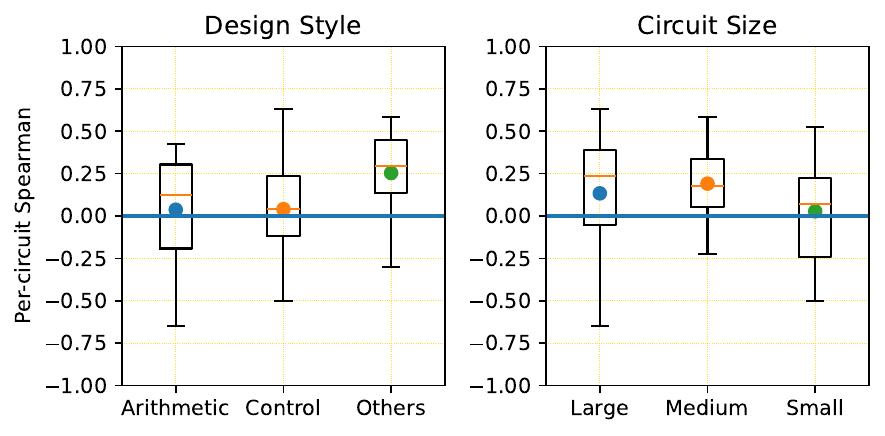} 
        \includegraphics[scale=0.6, trim = {0cm 0cm 0cm 0cm}, clip]{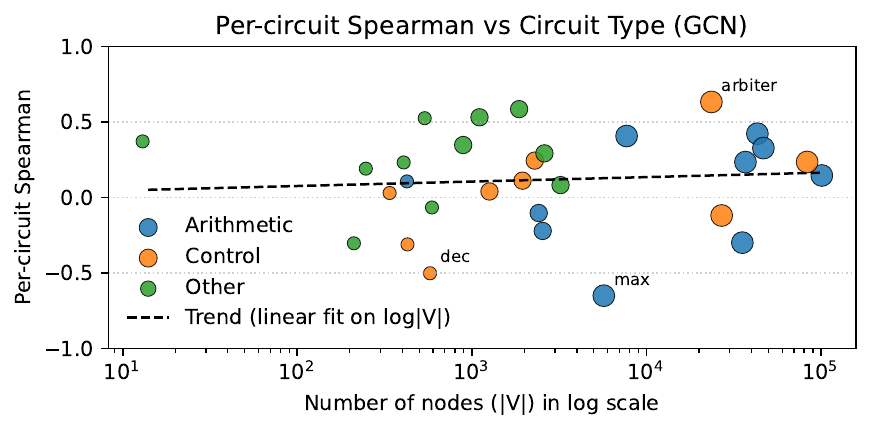}
		\caption{Robustness of learned structural manipulability across design styles and circuit sizes using a GCN. \emph{Arithmetic} benchmarks are datapath-intensive designs
(\texttt{adder.v}, \texttt{c6288.v}, \texttt{div.v}, \texttt{log2.v},
\texttt{max.v}, \texttt{multiplier.v}, \texttt{sin.v},
\texttt{sqrt.v}, \texttt{square.v}, \texttt{int2float.v}).
\emph{Control} benchmarks are controller or communication/memory
management circuits (\texttt{ctrl.v}, \texttt{memctrl.v}, \texttt{router.v},
\texttt{voter.v}, \texttt{arbiter.v}, \texttt{cavlc.v}, \texttt{dec.v},
\texttt{i2c.v}, \texttt{Priority.v}).
All remaining ISCAS/EPFL designs are grouped as \emph{Other}.
Circuit sizes are categorized into \emph{Small}, \emph{Medium}, and
\emph{Large} according to the number of nodes $|V|$ in each netlist.
The thresholds are determined by computing the 33rd and 66th percentile
(tertile) values of $|V|$ across all circuits of ISCAS85 and EPFL. Designs with
$|V| \le q_{1}$ are labeled \emph{Small}, those with
$q_{1} < |V| \le q_{2}$ are labeled \emph{Medium}, and those with
$|V| > q_{2}$ are labeled \emph{Large}, where $q_{1}$ and $q_{2}$
denote the first and second tertile cutoffs.} 
		\label{fig:robustness_style_and_size}
	\end{center}
    \vspace{-10pt}
\end{figure}

Fig.~\ref{fig:robustness_style_and_size} examines structural manipulability prediction across heterogeneous circuit types and scales. A representative GCN is trained on all circuits of ISCAS85 and EPFL using the composite target $M(v)$, and per-circuit Spearman correlation is computed between predicted and ground-truth $M(v)$ for each circuit. All reported per-circuit correlations are obtained from the trained GCN model. 

\textbf{Design-style behavior.} 
The left boxplot groups circuits by design style (Arithmetic, Control, Other). The distributions show that performance is not confined to a single design paradigm: the ``Other'' group attains the highest average correlation, while Arithmetic and Control exhibit lower means and larger dispersion, with notable negative outliers (e.g., \texttt{max.v}, \texttt{dec.v}). The bottom scatter plot further visualizes this effect by plotting per-circuit Spearman against circuit size ($|V|$ in log scale) and coloring points by design style, highlighting that both high and low correlations occur across all three categories. The presence of outliers and broad variance indicates that structural composition plays a significant role in prediction quality. 

\textbf{Size-related behavior.} 
The right boxplot groups circuits into Small/Medium/Large by node-count tertiles. Medium-sized circuits achieve the highest mean Spearman, whereas Small circuits show the lowest average correlation. The bottom scatter plot shows that despite substantial variation, the fitted trend over $\log(|V|)$ is weak, indicating that there is no strong monotonic scaling of ranking quality with circuit size. Large circuits can yield either strong positive correlations (e.g., \texttt{arbiter}) or negative outliers (e.g., \texttt{max}), suggesting that size alone does not determine prediction reliability. 

\textbf{Implication.} 
Overall, Fig.~\ref{fig:robustness_style_and_size} indicates partial robustness of structural manipulability prediction across design styles and circuit sizes, but with nontrivial variance and the presence of outliers. The weak dependence on size and the variability across design categories suggest that learnability is influenced more by structural composition than by simple graph-scale effects. These results should therefore be interpreted as indicative of conditional consistency under the evaluated benchmark setting, rather than uniform robustness across all circuit types.

\vspace{-5pt}
\subsection{Structural Manipulability in Trojan-Injected Circuits}
\label{subsec:trojan_analysis}

As an illustrative structural case study, we analyze Trojan-injected netlists derived from the same benchmark circuits. We adopt well-known templates from TrustHub ({\footnotesize\url{https://trust-hub.org/}}), namely \emph{CounterMUX}, \emph{FSMOR}, and \emph{ANDXOR}. These Trojans are synthetic insertion patterns designed for benchmarking purposes, and the analysis presented here is intended to examine structural distinguishability under these controlled settings rather than to establish a practical detection framework. Detailed experiment settings are given in \cite{gnnmutability2026}.

\begin{figure}[!t]
	\begin{center}
		\includegraphics[scale=0.6, trim = {0cm 0cm 0cm 0cm}, clip]{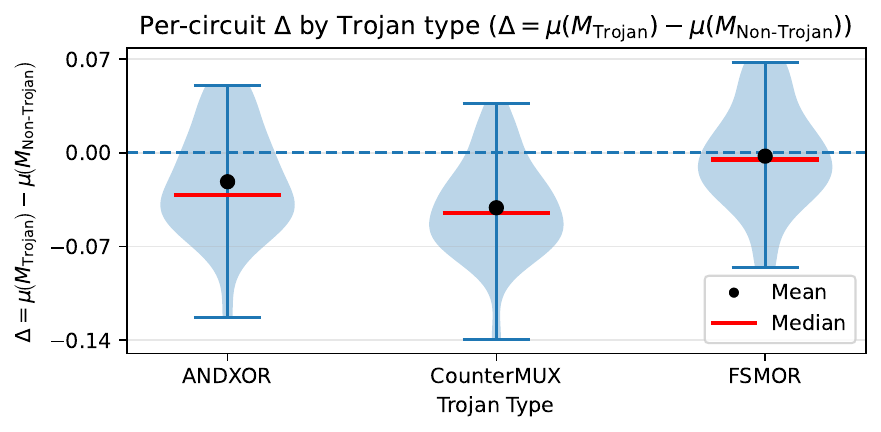} 
		\caption{Per-circuit distribution of 
$\Delta = \mu(M_{\text{Trojan}}) - \mu(M_{\text{Non-Trojan}})$ 
for three Trojans. \emph{CounterMUX:} It leverages a multiplexer whose select line activates only under rare trigger conditions. \emph{FSMOR:} This exploits finite state machines by inserting additional hidden states or transitions that cannot be reached during regular operation. \emph{ANDXOR:} This Trojan uses AND/XOR gates that remain dormant for most inputs. On rare and specific input patterns, the Trojan logic is enabled, resulting in corrupted computations. Each point on the plot represents one Trojanized benchmark circuit. Violin plots show the full distribution across circuits. } 
		\label{fig:trojan_delta_distribution}
	\end{center}
    \vspace{-10pt}
\end{figure}

For each Trojanized circuit, we compute the predicted structural manipulability $\hat{M}(v)$ for all gate nodes using the trained GNN model. We then compare the mean predicted manipulability of Trojan-inserted gates with that of non-Trojan gates. For each circuit, we define
\begin{equation}
    \Delta = \mu(M_{\text{Trojan}}) - \mu(M_{\text{Non-Trojan}}),
\end{equation}
where $\mu(M_{\text{Trojan}})$ denotes the mean predicted manipulability over Trojan gates and $\mu(M_{\text{Non-Trojan}})$ denotes the mean over all other gates in the same circuit. Positive $\Delta$ indicates Trojan regions are structurally more flexible, while negative $\Delta$ indicates they are more structurally embedded or critical. Table~\ref{tab:trojan_delta} summarizes the distribution of $\Delta$ across all circuits for each Trojan type, and Fig.~\ref{fig:trojan_delta_distribution} visualizes the per-circuit $\Delta$ distributions.

For ANDXOR Trojans, the mean $\Delta$ is $-0.0215$, with only $30\%$ of circuits exhibiting $\Delta>0$. The {\em Wilcoxon signed-rank} \cite{taheri2013generalization} test yields $p=0.0087$, indicating a statistically significant shift toward negative $\Delta$. CounterMUX Trojans exhibit an even stronger negative trend (mean $\Delta=-0.0410$, $p=2.0\times10^{-6}$), suggesting these Trojans are typically inserted in structurally embedded regions under these templates. In contrast, FSMOR Trojans show near-zero mean difference ($\Delta=-0.0024$) and no statistically significant deviation from zero ($p=0.6554$), indicating a more neutral structural placement pattern. 

Importantly, the absence of uniformly positive $\Delta$ indicates that the learned manipulability score is not trivially biased toward Trojan gates added during netlist modification. Instead, it reflects the actual structural embedding of those added gates within the surrounding circuit graph. 

These results indicate that, under the considered synthetic insertion models, Trojan gates can exhibit statistically distinguishable structural characteristics relative to the surrounding circuit. However, this analysis is limited to aggregate statistical comparison and does not establish a standalone methodology for Trojan detection, localization, or classification. Rather, it illustrates that the proposed topology-based structural score can capture differences in structural embedding under controlled insertion scenarios, providing complementary insight into netlist structure independent of functional simulation.

\begin{table}[!t]
\vspace{15pt}
\centering
\caption{Per-Trojan-type comparison of predicted structural manipulability. 
For each circuit, $\Delta = \mu(M_{\text{Trojan}}) - \mu(M_{\text{Non}})$ denotes the difference between the mean predicted manipulability of Trojan gates and non-Trojan gates.}
\label{tab:trojan_delta}
\footnotesize
\begin{tabular}{lcccc}
\toprule
\textbf{Trojan Type} & \textbf{Mean $\Delta$} & \textbf{Median $\Delta$} & \textbf{Frac($\Delta>0$)} & \textbf{Wilcoxon $p$} \\
\midrule
ANDXOR      & $-0.0215$ & $-0.0316$ & $0.300$ & $0.0087$ \\
CounterMUX  & $-0.0410$ & $-0.0448$ & $0.167$ & $2.0\times10^{-6}$ \\
FSMOR       & $-0.0024$ & $-0.0050$ & $0.433$ & $0.6554$ \\
\bottomrule
\end{tabular}
\vspace{-10pt}
\end{table}

\vspace{-5pt}
\section{Conclusion}

This work presented a topology-driven framework for analyzing structural manipulability in gate-level netlists by defining a composite graph-theoretic score derived from connectivity alone. Experiments on ISCAS85 and EPFL benchmarks showed that GNNs can approximate this topology-based score across held-out circuits, with architectural choices affecting ranking consistency. Component and ablation analyses indicated that core embedding dominates the metric, with centrality and symmetry providing secondary refinement. A preliminary case study on Trojan-injected circuits using synthetic TrustHub templates showed statistically distinguishable structural patterns under controlled settings. Overall, the results demonstrate that topology-based structural scoring can be learned using GNNs, while its practical applicability to downstream EDA or security tasks remains an open direction for future work.

\bibliographystyle{IEEEtran}
\bibliography{paper}

\appendix

\begin{table*}[!t]
\centering
\caption{Sensitivity of GNN prediction to alternative component weights. 
The weight schemes are uniform $(0.25,0.25,0.25,0.25)$, core-light $(0.30,0.10,0.30,0.30)$, and core-heavy $(0.20,0.40,0.20,0.20)$, where weights are ordered as $(M_{\text{path}},M_{\text{core}},M_{\text{sym}},M_{\text{cent}})$.}
\label{tab:weight_sensitivity_perf}
\small
\setlength{\tabcolsep}{3pt}
\begin{tabular}{llccc}
\toprule
\textbf{Model} & \textbf{Scheme} & \textbf{MSE} & \textbf{MAE} & \textbf{Spearman} \\
\midrule
\multirow{3}{*}{GCN}
& Uniform    & $0.000405 \pm 0.000076$ & $0.012738 \pm 0.001813$ & $0.6970 \pm 0.0254$ \\
& Core-light & $0.000094 \pm 0.000014$ & $0.007725 \pm 0.000451$ & $0.8139 \pm 0.0050$ \\
& Core-heavy & $0.000971 \pm 0.000188$ & $0.018450 \pm 0.003245$ & $0.6823 \pm 0.0158$ \\
\midrule
\multirow{3}{*}{GIN}
& Uniform    & $0.001222 \pm 0.001153$ & $0.016900 \pm 0.007062$ & $0.7784 \pm 0.0183$ \\
& Core-light & $0.000120 \pm 0.000045$ & $0.008011 \pm 0.000789$ & $0.7703 \pm 0.0225$ \\
& Core-heavy & $0.010760 \pm 0.013739$ & $0.036403 \pm 0.025431$ & $0.5614 \pm 0.1649$ \\
\midrule
\multirow{3}{*}{GSAGE}
& Uniform    & $0.000422 \pm 0.000083$ & $0.013435 \pm 0.001949$ & $0.7622 \pm 0.0292$ \\
& Core-light & $0.000092 \pm 0.000016$ & $0.007765 \pm 0.000635$ & $0.7810 \pm 0.0438$ \\
& Core-heavy & $0.001060 \pm 0.000229$ & $0.019744 \pm 0.003278$ & $0.7538 \pm 0.0115$ \\
\midrule
\multirow{3}{*}{GAT}
& Uniform    & $0.000414 \pm 0.000184$ & $0.012791 \pm 0.002909$ & $0.6927 \pm 0.0259$ \\
& Core-light & $0.000092 \pm 0.000029$ & $0.007633 \pm 0.000880$ & $0.7394 \pm 0.0242$ \\
& Core-heavy & $0.000998 \pm 0.000501$ & $0.018119 \pm 0.005557$ & $0.6861 \pm 0.0330$ \\
\midrule
\multirow{3}{*}{MPNN}
& Uniform    & $0.000449 \pm 0.000110$ & $0.013835 \pm 0.001945$ & $0.7495 \pm 0.0258$ \\
& Core-light & $0.000100 \pm 0.000019$ & $0.007855 \pm 0.000302$ & $0.7430 \pm 0.0066$ \\
& Core-heavy & $0.001144 \pm 0.000297$ & $0.020454 \pm 0.003564$ & $0.7247 \pm 0.0136$ \\
\midrule
\multirow{3}{*}{APPNP}
& Uniform    & $0.000413 \pm 0.000086$ & $0.012548 \pm 0.001464$ & $0.7266 \pm 0.0213$ \\
& Core-light & $0.000094 \pm 0.000016$ & $0.007693 \pm 0.000378$ & $0.7945 \pm 0.0463$ \\
& Core-heavy & $0.001001 \pm 0.000208$ & $0.018149 \pm 0.002751$ & $0.7148 \pm 0.0114$ \\
\midrule
\multirow{3}{*}{G-U-Net}
& Uniform    & $0.000404 \pm 0.000124$ & $0.013040 \pm 0.001892$ & $0.7327 \pm 0.0287$ \\
& Core-light & $0.000094 \pm 0.000020$ & $0.007793 \pm 0.000521$ & $0.7617 \pm 0.0132$ \\
& Core-heavy & $0.000958 \pm 0.000301$ & $0.018682 \pm 0.003057$ & $0.7056 \pm 0.0174$ \\
\midrule
\multirow{3}{*}{HetGNN}
& Uniform    & $0.000406 \pm 0.000135$ & $0.012079 \pm 0.002675$ & $0.8036 \pm 0.0155$ \\
& Core-light & $0.000088 \pm 0.000019$ & $0.007427 \pm 0.000726$ & $0.8014 \pm 0.0476$ \\
& Core-heavy & $0.000968 \pm 0.000343$ & $0.017178 \pm 0.004490$ & $0.7915 \pm 0.0023$ \\
\midrule
\multirow{3}{*}{SGNN}
& Uniform    & $0.000467 \pm 0.000057$ & $0.013251 \pm 0.001875$ & $0.7623 \pm 0.0408$ \\
& Core-light & $0.000096 \pm 0.000012$ & $0.007663 \pm 0.000727$ & $0.7772 \pm 0.0425$ \\
& Core-heavy & $0.001127 \pm 0.000130$ & $0.019239 \pm 0.003554$ & $0.7470 \pm 0.0421$ \\
\midrule
\multirow{3}{*}{GTN}
& Uniform    & $0.000457 \pm 0.000106$ & $0.013205 \pm 0.001159$ & $0.7433 \pm 0.0374$ \\
& Core-light & $0.000098 \pm 0.000011$ & $0.007743 \pm 0.000093$ & $0.7567 \pm 0.0355$ \\
& Core-heavy & $0.001068 \pm 0.000194$ & $0.018970 \pm 0.002438$ & $0.7599 \pm 0.0636$ \\
\bottomrule
\end{tabular}
\end{table*}

\begin{table*}[!t]
\centering
\vspace{10pt}
\caption{Structural composition of weighted targets under different component-weight schemes. 
$\rho_{\text{core}}=\rho(M_{\mathbf{w}},M_{\text{core}})$, 
$\rho_{\text{sym}}=\rho(M_{\mathbf{w}},M_{\text{sym}})$, and 
$\rho_{\text{flow}}=\rho(M_{\mathbf{w}},M_{\text{path}}+M_{\text{cent}})$.}
\label{tab:weight_sensitivity_comp}
\small
\setlength{\tabcolsep}{3pt}
\begin{tabular}{lcccc}
\toprule
\textbf{Scheme} & $\boldsymbol{\rho_{\text{core}}}$ & $\boldsymbol{\rho_{\text{sym}}}$ & $\boldsymbol{\rho_{\text{flow}}}$ & $\boldsymbol{\sigma(M_{\mathbf{w}})}$ \\
\midrule
Uniform    & $0.8405 \pm 0.0317$ & $0.5169 \pm 0.1084$ & $0.1700 \pm 0.2228$ & $0.0413 \pm 0.0011$ \\
Core-light & $0.8396 \pm 0.0312$ & $0.5178 \pm 0.1082$ & $0.1705 \pm 0.2217$ & $0.0180 \pm 0.0002$ \\
Core-heavy & $0.8405 \pm 0.0317$ & $0.5169 \pm 0.1084$ & $0.1699 \pm 0.2228$ & $0.0653 \pm 0.0021$ \\
\bottomrule
\end{tabular}
\vspace{-8pt}
\end{table*}

\subsection{Sensitivity to Component Weights}
\label{subsec:weight_sensitivity}

To examine whether the observed component imbalance is tied to the uniform weighting choice, we conduct a weight-sensitivity experiment using alternative definitions of the composite score $M_{\mathbf{w}}(v)=\alpha M_{\text{path}}(v)+\beta M_{\text{core}}(v)+\gamma M_{\text{sym}}(v)+\delta M_{\text{cent}}(v)$, where the weights are ordered as $(\alpha,\beta,\gamma,\delta)$. In addition to the uniform setting $(0.25,0.25,0.25,0.25)$, we evaluate a \emph{core-light} setting $(0.30,0.10,0.30,0.30)$, which explicitly reduces the contribution of $M_{\text{core}}$, and a \emph{core-heavy} setting $(0.20,0.40,0.20,0.20)$, which increases the influence of $M_{\text{core}}$. These settings are not intended to define optimal weights; rather, they test whether the proposed formulation remains learnable when the relative emphasis on core structure is varied. Since all tested schemes use equal weights for the two flow-related terms, we report their combined target correlation as $\rho(M_{\mathbf{w}},M_{\text{path}}+M_{\text{cent}})$.

Table~\ref{tab:weight_sensitivity_perf} shows that the core-light setting generally preserves or improves rank consistency across most architectures, while core-heavy weighting often increases regression error and reduces Spearman correlation. Since the target variance differs across weighting schemes, MSE and MAE should be interpreted together with the rank correlation. The target-composition statistics in Table~\ref{tab:weight_sensitivity_comp} show that all schemes remain strongly correlated with $M_{\text{core}}$, indicating that core dominance is partly induced by the benchmark component distributions rather than by uniform weights alone. These results support using uniform weights as a transparent baseline, while also showing that non-uniform, application-specific weights can be adopted without changing the proposed framework.

The uniform rows in Table~\ref{tab:weight_sensitivity_perf} should not be interpreted as duplicates of Table~\ref{tab:gnn_comparison}. Table~\ref{tab:gnn_comparison} reports the main architectural comparison under the original experimental protocol, where the uniformly weighted metric is used as the primary target; for example, GCN achieves Spearman $0.7881 \pm 0.0126$ and G-U-Net achieves $0.8129 \pm 0.0205$. In contrast, Table~\ref{tab:weight_sensitivity_perf} re-runs the models under a controlled weight-sensitivity protocol so that the uniform, core-light, and core-heavy targets are compared under the same splits and training setup. Thus, the uniform row in Table~\ref{tab:weight_sensitivity_perf} serves as an internal reference for measuring the effect of changing weights, rather than replacing the main results in Table~\ref{tab:gnn_comparison}. Under this controlled setting, reducing the core weight often preserves or improves ranking consistency, e.g., GCN improves from $0.6970 \pm 0.0254$ to $0.8139 \pm 0.0050$, GSAGE from $0.7622 \pm 0.0292$ to $0.7810 \pm 0.0438$, and APPNP from $0.7266 \pm 0.0213$ to $0.7945 \pm 0.0463$ under the core-light setting. Conversely, increasing the core weight can degrade learnability, as seen for GIN, where Spearman decreases from $0.7784 \pm 0.0183$ to $0.5614 \pm 0.1649$ under core-heavy weighting. These trends suggest that reducing $\beta$ can mitigate excessive emphasis on $M_{\text{core}}$ while keeping the target learnable, although Table~\ref{tab:weight_sensitivity_comp} shows that the target remains strongly correlated with $M_{\text{core}}$ across all schemes, indicating that core dominance is also driven by the benchmark component distributions.

\subsection{Potential Use Cases of Structural Manipulability}
\label{subsec:use_cases}

The proposed score $M(v)$ is intended as a topology-based structural characterization primitive rather than a replacement for task-specific EDA or security tools. In Section~\ref{subsec:trojan_analysis}, we already demonstrate one security-oriented use case by showing that Trojan-inserted regions can exhibit statistically distinguishable structural characteristics under synthetic TrustHub-style insertion models. Beyond this demonstrated case, the same node-level score can support several practical workflows as an auxiliary ranking, screening, or prioritization signal.

\subsubsection{Hardware Trojan Triage and Security Screening}
Hardware Trojan analysis often requires identifying suspicious regions in a netlist before applying more expensive functional, formal, or side-channel checks \cite{salmani2013trusthub,yasaei2022hardware}. The proposed score can help by ranking nodes according to structural embedding, bottleneck participation, and local redundancy. Regions with unusual $M(v)$ distributions relative to the surrounding circuit can be prioritized for additional Trojan detection or localization analysis. This use case should be viewed as auxiliary triage rather than a standalone Trojan detector.

\subsubsection{Engineering Change Order and Edit-Impact Estimation}
Engineering Change Order (ECO) flows aim to implement late-stage design changes with minimal disruption to an existing implementation \cite{jiang2020eco}. Nodes with higher $M(v)$ can be interpreted as structurally more flexible candidates for localized modification, while low-$M(v)$ nodes may indicate structurally critical regions where edits could propagate broadly. Thus, $M(v)$ can guide ECO engineers toward regions where patch insertion or local resynthesis may be less structurally disruptive. It can also be used to flag low-manipulability regions that require more careful timing, equivalence, or regression validation.

\subsubsection{Design-for-Test and Test-Point Insertion}
Test-point insertion improves controllability and observability by adding control or observation points to selected circuit locations \cite{touba1996testpoint,shi2022deeptpi}. Since $M(v)$ combines path participation, centrality, and structural redundancy, it can help identify nodes whose modification may improve structural access without targeting only obvious high-fanout locations. High-$M(v)$ nodes may be useful candidates for low-disruption observation or control insertion, whereas low-$M(v)$ nodes may correspond to bottlenecks whose instrumentation could affect many paths. Therefore, the score can serve as a structural pre-filter before ATPG- or simulation-driven test-point selection.

\subsubsection{Logic Locking and Obfuscation Placement}
Logic locking inserts key-controlled elements to protect IP against unauthorized use or overproduction \cite{dupuis2019logiclocking}. The placement of locking elements affects both security and design overhead, and structural position is one factor in this tradeoff. $M(v)$ can help distinguish flexible regions suitable for low-overhead key-gate insertion from highly central regions where insertion may cause larger structural or timing impact. Conversely, deliberately selecting lower-$M(v)$ bottleneck regions may increase functional influence, but would require stronger validation to avoid excessive overhead.

\subsubsection{Reliability and Soft-Error Hardening}
Soft-error analysis studies whether transient faults at internal nodes can propagate to observable outputs \cite{rao2007ser,miskov2006marsc}. Nodes with low $M(v)$ often participate more strongly in global paths or bottleneck structures, making them natural candidates for reliability screening. The score can therefore help prioritize gates for selective hardening, redundancy insertion, or fault-injection simulation. This does not replace electrical or timing-aware SER estimation, but can reduce the search space for detailed reliability analysis.

\subsubsection{Synthesis, Placement, and Congestion-Aware Structural Screening}
Modern EDA flows increasingly use graph-based models to predict physical or structural properties such as timing, congestion, or optimization difficulty \cite{zhao2023hybridnet,dong2023cktgnn}. Since $M(v)$ summarizes structural embedding and connectivity at node level, it can serve as an additional graph feature for downstream prediction tasks. Regions with low manipulability may correspond to structurally constrained subgraphs where optimization, placement, or buffering decisions require more care. Regions with high manipulability may provide candidate areas for local restructuring or low-risk optimization.

\begin{table*}[!t]
\centering
\caption{Comparison of the composite score $M(v)$ with individual structural-factor targets using a representative GCN. 
Here, $T(v)$ denotes the target being predicted, $\rho(T,M)$ is the Spearman correlation between $T(v)$ and the composite score $M(v)$, and top-$10\%$ overlap measures the fraction of highest-scoring nodes shared with $M(v)$. Results are reported as mean $\pm$ std over two circuit-level splits.}
\label{tab:single_component_comparison}
\footnotesize
\setlength{\tabcolsep}{4pt}
\begin{tabular}{lcccccc}
\toprule
\textbf{Target $T(v)$} & $\boldsymbol{\rho(T,M)}$ & \textbf{Top-$10\%$ overlap} & $\boldsymbol{\sigma(T)}$ & \textbf{MSE} & \textbf{MAE} & \textbf{Spearman} \\
\midrule
Composite $M(v)$ 
& $1.0000 \pm 0.0000$ 
& $1.0000 \pm 0.0000$ 
& $0.0223 \pm 0.0005$ 
& $0.000405 \pm 0.000076$ 
& $0.012738 \pm 0.001812$ 
& $0.6970 \pm 0.0253$ \\

Core-only $M_{\text{core}}(v)$ 
& $0.8405 \pm 0.0317$ 
& $0.2908 \pm 0.1006$ 
& $0.0787 \pm 0.0148$ 
& $0.021444 \pm 0.006179$ 
& $0.097924 \pm 0.026005$ 
& $0.8270 \pm 0.0251$ \\

Sym-only $M_{\text{sym}}(v)$ 
& $0.5169 \pm 0.1084$ 
& $0.4725 \pm 0.1276$ 
& $0.0242 \pm 0.0063$ 
& $0.000391 \pm 0.000031$ 
& $0.015313 \pm 0.000373$ 
& $0.5600 \pm 0.0083$ \\

Flow-only $M_{\text{flow}}(v)$ 
& $0.1700 \pm 0.2228$ 
& $0.2550 \pm 0.0174$ 
& $0.0053 \pm 0.0023$ 
& $0.000002 \pm 0.000002$ 
& $0.000242 \pm 0.000037$ 
& $0.3647 \pm 0.0443$ \\
\bottomrule
\end{tabular}
\end{table*}

\subsection{Composite Score Versus Individual Structural Factors}
\label{subsec:single_component_comparison}

To clarify how the proposed composite score differs from individual structural metrics, we compare $M(v)$ against single-factor targets using a representative GCN. Specifically, we evaluate the original composite score, a core-only target $M_{\text{core}}(v)$, a symmetry-only target $M_{\text{sym}}(v)$, and a flow-only target $M_{\text{flow}}(v)=\frac{1}{2}(M_{\text{path}}(v)+M_{\text{cent}}(v))$. The flow-only target combines the two global flow-related terms because $M_{\text{path}}$ and $M_{\text{cent}}$ both characterize path or bottleneck participation. For each target $T(v)$, Table~\ref{tab:single_component_comparison} reports its Spearman correlation with the composite score $M(v)$, top-$10\%$ overlap with the highest-scoring nodes under $M(v)$, target standard deviation, and GCN prediction performance. This experiment does not aim to show that the composite score is universally superior to every individual component; rather, it tests whether $M(v)$ is reducible to a single structural factor or provides a distinct multi-factor ranking.

Table~\ref{tab:single_component_comparison} shows that $M_{\text{core}}(v)$ is strongly correlated with the composite score ($\rho=0.8405$), confirming the core dominance observed earlier. However, the top-$10\%$ overlap between $M_{\text{core}}(v)$ and $M(v)$ is only $0.2908$, indicating that the highest-ranked nodes selected by the composite score are not simply the same as those selected by core structure alone. Symmetry has moderate correlation with $M(v)$ but lower learnability, while the flow-only target has very low variance and weak correlation with $M(v)$, explaining its limited standalone discriminative role. The core-only target is easier for GCN to rank (Spearman $0.8270$), but it represents a narrower structural view; in contrast, $M(v)$ provides an integrated ranking that combines embedding, symmetry, and flow-related signals. Thus, the proposed metric should be interpreted not as replacing individual graph metrics, but as a unified structural descriptor whose ranking is influenced by, yet not reducible to, any single component.

\begin{table}[!t]
\centering
\caption{Preprocessing, normalization, and training protocol used in the experiments.}
\label{tab:training_protocol}
\footnotesize
\setlength{\tabcolsep}{3pt}
\begin{tabular}{|l|p{0.62\columnwidth}|}
\toprule
\textbf{Item} & \textbf{Setting} \\
\midrule
Graph nodes & Gate instances plus PI/PO boundary nodes. Sequential cells, if present, are treated as typed nodes. \\ \hline
Graph edges & Directed producer-to-consumer signal edges are parsed; PyG message passing uses bidirectional edges. \\ \hline
Cycles & Cycles are preserved; no DAG assumption or temporal unrolling is used. \\ \hline
Node features $X_v$ & Global gate-type one-hot, PI indicator, PO indicator, fan-in, fan-out. \\ \hline
Training mask & Loss and metrics are computed only on gate nodes using \texttt{gate\_mask}. \\ \hline
Target normalization & Per-circuit normalization over gate nodes: $(M(v)-\mu_G)/\sigma_G$, with $\sigma_G=1$ if $\sigma_G<10^{-6}$. \\ \hline
Splitting & Circuit-level train/test splits; held-out circuits are not used during training. \\ \hline
Optimizer & Adam. \\ \hline
Learning rate & $10^{-3}$. \\ \hline
Loss & Mean squared error on normalized gate-node targets. \\ \hline
Batch size & 4 circuit graphs. \\ \hline
Hidden dimension & 64. \\ \hline
Depth & 3 message-passing layers unless otherwise required by the architecture. \\ \hline
Regularization & No explicit dropout or weight decay in the main protocol. \\ \hline
Early stopping & Not used; models are trained for a fixed epoch budget. \\ \hline
Epochs & 30 epochs for multi-seed evaluation; single-run sanity checks use up to 50 epochs. \\ \hline
Evaluation & MSE, MAE, and Spearman on unnormalized gate-node predictions. \\
\bottomrule
\end{tabular}
\vspace{-8pt}
\end{table}

\subsection{Preprocessing and Training Protocol}
\label{subsec:preprocess_training}

Gate-level netlists are parsed into graph objects before training. The parser first constructs a directed graph where each gate instance is represented as a node and each edge follows signal propagation from a producer gate or PI to a consumer gate or PO. PI and PO nodes are retained in the graph to preserve boundary connectivity, but the regression loss and reported node-level metrics are computed only over gate nodes using a Boolean gate mask. For message passing, each directed edge is converted into two directed edges in the PyG representation, making aggregation effectively bidirectional; the original fan-in and fan-out information is retained as node features. The node feature vector $X_v$ contains a one-hot encoding of the gate type using a global gate-type vocabulary, binary indicators for PI and PO nodes, and scalar fan-in/fan-out values. Sequential elements, when present as cell instances in the gate-level netlist, are treated as typed nodes in the same graph representation; no temporal unrolling or state-transition modeling is performed. Feedback cycles are not removed, and graph-theoretic quantities such as $k$-core, WL symmetry, and betweenness-based terms are computed on the undirected projection of the graph, so the preprocessing does not require the netlist graph to be acyclic.

For training, the target $M(v)$ is normalized independently per circuit using only gate nodes, i.e., $y_{\text{norm}}=(M(v)-\mu_G)/\sigma_G$, where $\mu_G$ and $\sigma_G$ are the mean and standard deviation of $M(v)$ over gate nodes in graph $G$; if $\sigma_G<10^{-6}$, we set $\sigma_G=1$. This per-circuit normalization is used only for numerical conditioning during training, and predictions are mapped back to the original $M(v)$ scale for MSE and MAE evaluation. Spearman correlation is computed on gate nodes and is invariant to this affine normalization. Train/test splits are performed at the circuit level, so all nodes from a held-out circuit are unseen during training. We use a fixed training budget rather than hyperparameter search or validation-based early stopping; this keeps architectural comparisons consistent and avoids tuning to a particular benchmark split. Table~\ref{tab:training_protocol} summarizes the preprocessing and training settings.

\begin{table*}[!t]
\centering
\caption{Comparison of GNNs with simple non-GNN regressors and structural rankers. 
Here, $\checkmark$ denotes that the attribute is present, $\times$ denotes absent, and P denotes partial support.}
\label{tab:non_gnn_baseline_attributes}
\footnotesize
\setlength{\tabcolsep}{4pt}
\begin{tabular}{|l|cccccc|p{0.23\textwidth}|}
\toprule
\textbf{Method} &
\textbf{Learned} &
\textbf{Uses $X_v$} &
\textbf{Uses edges} &
\textbf{Multi-hop topology} &
\textbf{Nonlinear} &
\textbf{Outputs rank} &
\textbf{Interpretation} \\
\midrule
Mean predictor 
& $\times$ & $\times$ & $\times$ & $\times$ & $\times$ & $\times$ 
& Predicts a constant score; sanity baseline for regression error. \\ \hline

Fan-in/fan-out ranker 
& $\times$ & P & $\times$ & $\times$ & $\times$ & $\checkmark$ 
& Uses local degree statistics only; cannot capture global embedding or bottlenecks. \\ \hline

Inverse betweenness ranker 
& $\times$ & $\times$ & $\checkmark$ & $\checkmark$ & $\times$ & $\checkmark$ 
& Uses global shortest-path structure; captures bottleneck avoidance similar to $M_{\text{cent}}$. \\ \hline

$k$-core ranker 
& $\times$ & $\times$ & $\checkmark$ & $\checkmark$ & $\times$ & $\checkmark$ 
& Captures dense subgraph embedding; useful for testing whether $M(v)$ is reducible to $M_{\text{core}}$. \\ \hline

Ridge regression 
& $\checkmark$ & $\checkmark$ & $\times$ & $\times$ & $\times$ & $\checkmark$ 
& Linear feature-based regressor; tests whether local node features alone explain $M(v)$. \\ \hline

Random forest 
& $\checkmark$ & $\checkmark$ & $\times$ & $\times$ & $\checkmark$ & $\checkmark$ 
& Nonlinear feature-based regressor; captures feature interactions but not graph message passing. \\ \hline

Gradient-boosted trees 
& $\checkmark$ & $\checkmark$ & $\times$ & $\times$ & $\checkmark$ & $\checkmark$ 
& Strong tabular regressor; useful for testing non-GNN nonlinear prediction from local features. \\ \hline

GNN surrogate 
& $\checkmark$ & $\checkmark$ & $\checkmark$ & $\checkmark$ & $\checkmark$ & $\checkmark$ 
& Uses node features and graph connectivity through message passing to approximate $M(v)$. \\
\bottomrule
\end{tabular}
\vspace{-8pt}
\end{table*}

\subsection{Comparison with Non-GNN Baselines}
\label{subsec:non_gnn_baselines}

To clarify the added role of graph neural networks, we compare the proposed GNN-based surrogate with simple non-GNN regressors and hand-crafted structural rankers. The goal of this comparison is not to claim that GNNs replace direct graph analysis, but to distinguish what information each class of method can exploit. We consider two categories of baselines. The first category consists of non-learned rankers derived from individual structural statistics, such as fan-in/fan-out degree, inverse betweenness centrality, and normalized $k$-core score \cite{freeman1977centrality,kitsak2010identification}. The second category consists of feature-based regressors, such as ridge regression, random forests, and gradient-boosted trees, which learn from node features but do not perform graph message passing \cite{hoerl1970ridge,breiman2001random,friedman2001greedy}. All learned baselines use the same circuit-level splits, gate-node mask, and target normalization as the GNN experiments.

Table~\ref{tab:non_gnn_baseline_attributes} summarizes the main differences among these baselines. Direct rankers are transparent and inexpensive, but each captures only one structural view. Feature-based regressors can learn nonlinear mappings from local node features, but they do not explicitly aggregate multi-hop neighborhood information. In contrast, GNNs combine node features with graph connectivity through message passing, enabling them to approximate structural quantities that depend on both local and global topology. This comparison helps separate the contribution of the proposed metric from the contribution of the learning architecture: simple rankers test whether one structural statistic is already sufficient, while non-GNN regressors test whether message passing provides additional value beyond local node features.

\subsection{Graph Direction Handling}
\label{subsec:direction_handling}

The gate-level netlist is first parsed as a directed graph, where edges follow signal propagation from producer nodes to consumer nodes. Thus, direction is used to identify PI/PO connectivity, fan-in, fan-out, and the original circuit topology. However, for GNN message passing, each directed edge is converted into two directed edges in the PyG representation, so aggregation is effectively bidirectional for all evaluated architectures. We do not use separate in-/out-adjacency matrices or explicit directional edge types. Instead, direction-related information is retained through node features, including PI/PO indicators and scalar fan-in/fan-out values. SGNN also uses the same bidirectional netlist edge set as its positive connectivity; its signed-edge construction does not encode circuit signal direction.

This design choice is intentional because the prediction target $M(v)$ is a structural-topology score rather than a timing-accurate signal-propagation model. Several components of $M(v)$, such as $k$-core, WL symmetry, and betweenness-based quantities, are computed on the undirected projection of the parsed netlist graph, while still using directed information to identify boundary nodes and degree features. Bidirectional propagation therefore aligns the learning model with the structural quantities being predicted and allows information from both fan-in and fan-out neighborhoods to contribute to node embeddings. We did not separately ablate directed-only versus bidirectional propagation; therefore, the reported results should be interpreted as evaluating structural, direction-aware-in-features but bidirectional-in-aggregation message passing, rather than a fully directed GNN formulation.

\subsection{Interpretation of Key Experimental Results}
\label{subsec:result_interpretation}

\textbf{Table~\ref{tab:gnn_comparison}.}
Table~\ref{tab:gnn_comparison} compares how different GNN architectures approximate the topology-derived score $M(v)$. Most message-passing models achieve low regression error and positive rank correlation, indicating that $M(v)$ is learnable from local graph aggregation. However, the variation across models also shows that architectural bias matters: G-U-Net and GTN provide stronger ranking consistency, while SGNN underperforms, suggesting that signed-edge assumptions are not naturally aligned with the netlist graphs used here. Thus, the table supports learnability of the proposed structural score, but not uniform performance across all GNN designs.

\textbf{Table~\ref{tab:component_stats}.}
Table~\ref{tab:component_stats} explains why the composite score is not empirically balanced across components. The strongest association with $M(v)$ comes from $M_{\text{core}}$, while $M_{\text{path}}$ has high mean and low variance, making it weakly discriminative in these benchmarks. This means that the composite score behaves as a core-centered structural descriptor enriched by symmetry and centrality, rather than as an equal empirical mixture of all four components. This interpretation is important because the weights define the metric, but the benchmark distributions determine how strongly each component affects node ranking.

\textbf{Fig.~\ref{fig:per_circuit_spearman}.}
Fig.~\ref{fig:per_circuit_spearman} shows that aggregate performance can hide substantial circuit-level variability. Some circuits yield positive rank correlation across models, while others produce negative Spearman values, meaning that predicted rankings can be partially inverted relative to the ground-truth structural score. This indicates that the learned surrogate is sensitive to circuit structure and does not provide uniformly reliable rankings across all held-out designs. Therefore, the figure should be interpreted as evidence of conditional cross-circuit transfer rather than universal generalization.

\textbf{Table~\ref{tab:per_circuit_stats}.}
Table~\ref{tab:per_circuit_stats} quantifies the trends visible in Fig.~\ref{fig:per_circuit_spearman}. The family-wise means show that ISCAS and EPFL circuits can behave differently under the same model and seed, while $\rho(|V|,\text{Spr})$ captures how performance changes with graph size. The $\Delta_{\text{wt-unwt}}$ column further shows whether node-weighted aggregate performance is dominated by larger circuits. Thus, this table clarifies that benchmark family and circuit size can influence the reported global metrics.

\textbf{Table~\ref{tab:ablation}.}
Table~\ref{tab:ablation} identifies which components most affect learnability of the composite score. Removing $M_{\text{core}}$ causes the largest drop in Spearman correlation, confirming that dense structural embedding is the main ranking driver. Removing $M_{\text{cent}}$ also degrades performance, indicating that global flow and bottleneck structure contribute meaningful information. In contrast, removing $M_{\text{sym}}$ has little effect on rank correlation, suggesting that symmetry provides secondary refinement rather than the dominant signal in these benchmarks.

\textbf{Table~\ref{tab:global_stat_correlation}.}
Table~\ref{tab:global_stat_correlation} examines whether simple circuit-level statistics explain prediction quality. The weak negative correlations for node and edge counts indicate that larger circuits are not automatically easier to predict. The moderate positive correlations for average fan-in and fan-out suggest that local connectivity density provides stronger structural cues than graph size alone. However, all correlations are modest, so these global statistics only partially explain per-circuit variability.

\textbf{Fig.~\ref{fig:robustness_style_and_size}.}
Fig.~\ref{fig:robustness_style_and_size} visualizes how prediction quality varies across design styles and size groups. The design-style boxplots show that the ``Other'' category has higher average performance, while Arithmetic and Control circuits show larger dispersion and negative outliers. The size plot and scatter trend show no strong monotonic relationship between $|V|$ and Spearman correlation: large circuits such as \texttt{arbiter} can be predicted well, while other large designs such as \texttt{max} can be outliers. This supports the conclusion that structural composition, not size alone, governs prediction reliability.

\textbf{Table~\ref{tab:trojan_delta}.}
Table~\ref{tab:trojan_delta} summarizes how the predicted structural score differs between Trojan and non-Trojan gates using $\Delta=\mu(M_{\text{Trojan}})-\mu(M_{\text{Non-Trojan}})$. Negative mean and median $\Delta$ values for ANDXOR and CounterMUX indicate that these inserted gates tend to lie in more structurally embedded or critical regions under the proposed score. The Wilcoxon $p$-values show that these negative shifts are statistically significant for ANDXOR and CounterMUX, but not for FSMOR. Therefore, the table indicates template-dependent structural distinguishability, not a general Trojan detection capability.

\textbf{Fig.~\ref{fig:trojan_delta_distribution}.}
Fig.~\ref{fig:trojan_delta_distribution} complements Table~\ref{tab:trojan_delta} by showing the per-circuit distribution of $\Delta$ for each Trojan type. The violin shapes reveal that CounterMUX is consistently shifted toward negative $\Delta$, ANDXOR is also mostly negative but more dispersed, and FSMOR is centered closer to zero. The mean and median markers help distinguish systematic shifts from isolated outliers. Since the distributions are not uniformly positive, the result also shows that the learned score is not simply biased toward inserted Trojan gates; instead, it reflects how the inserted gates are structurally embedded in each benchmark circuit.

\end{document}